\documentclass[10pt,twocolumn,letterpaper]{article}

\usepackage{cvpr}
\usepackage{times}
\usepackage{epsfig}
\usepackage{graphicx}
\usepackage{amsmath}
\usepackage{amssymb}
\usepackage{booktabs}
\usepackage{capt-of,etoolbox}

\newcommand\blfootnote[1]{%
  \begingroup
  \renewcommand\thefootnote{}\footnote{#1}%
  \addtocounter{footnote}{-1}%
  \endgroup
}

\usepackage[pagebackref=true,breaklinks=true,letterpaper=true,colorlinks,bookmarks=false]{hyperref}

\cvprfinalcopy 

\def\cvprPaperID{3523} 

\ifcvprfinal\fi
\begin{document}

\title{TextureGAN: Controlling Deep Image Synthesis with Texture Patches}

\author{Wenqi Xian $^{\dagger 1}$ \and Patsorn Sangkloy $^{\dagger 1}$ \and Varun Agrawal $^1$ \and Amit  Raj $^1$ \and Jingwan Lu $^2$ \quad  Chen Fang $^2$ \quad Fisher Yu $^3$ \quad  James Hays $^{1,4}$ 
\vspace{1.5mm}\\
$^1$Georgia Institute of Technology \quad $^2 $Adobe Research \quad $^3 $UC Berkeley \quad $^4$Argo AI\\
\vspace{0mm}
}

\twocolumn[{%
\renewcommand\twocolumn[1][]{#1}%
\maketitle
\vspace*{-4em}
\begin{center}
\centering
\includegraphics[width=\textwidth]{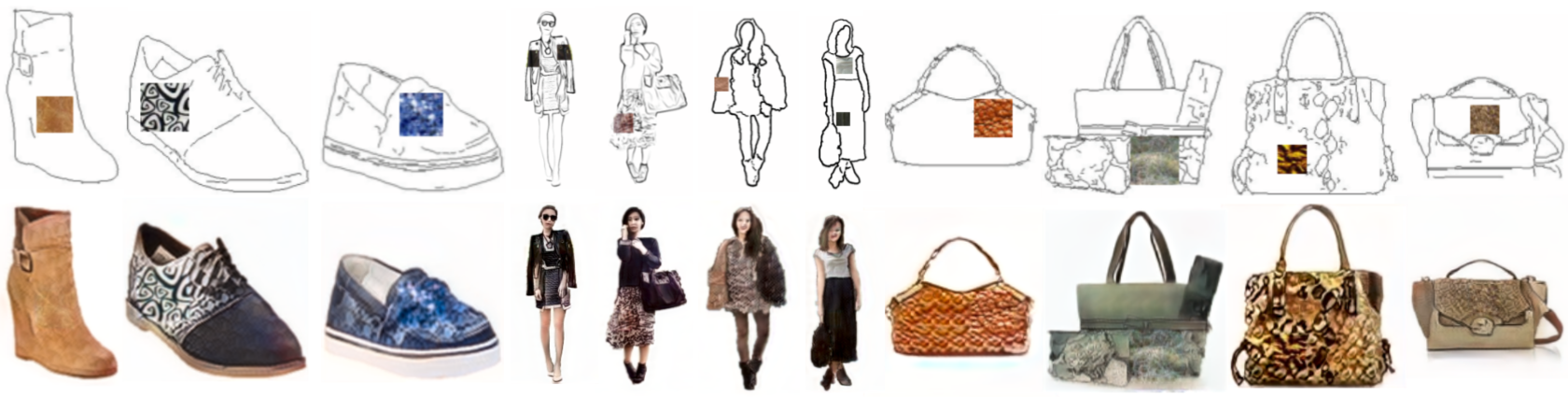}
\captionof{figure}{With TextureGAN, one can generate novel instances of common items from hand drawn sketches and simple texture patches. You can now be your own fashion guru! Top row: Sketch with texture patch overlaid. Bottom row: Results from TextureGAN.}
\label{fig:teaser}
\end{center}
}]


\begin{abstract}
In this paper, we investigate deep image synthesis guided by sketch, color, and \emph{texture}. Previous image synthesis methods can be controlled by sketch and color strokes but we are the first to examine texture control. We allow a user to place a texture patch on a sketch at arbitrary locations and scales to control the desired output texture.  Our generative network learns to synthesize objects consistent with these texture suggestions. To achieve this, we develop a local texture loss in addition to adversarial and content loss to train the generative network. We conduct experiments using sketches generated from real images and textures sampled from a separate texture database
and results show that our proposed algorithm is able to generate plausible images that are faithful to user controls. Ablation studies show that our proposed pipeline can generate more realistic images than adapting existing methods directly.
\end{abstract}

\blfootnote{$^\dagger$ indicates equal contribution}

\section{Introduction}

One of the ``Grand Challenges'' of computer graphics is to allow \emph{anyone} to author realistic visual content. The traditional 3d rendering pipeline can produce astonishing and realistic imagery, but only in the hands of talented and trained artists. The idea of short-circuiting the traditional 3d modeling and rendering pipeline dates back at least 20 years to image-based rendering techniques~\cite{mcmillan1995plenoptic}. These techniques and later ``image-based'' graphics approaches focus on re-using image content from a database of training images~\cite{lalonde-siggraph-07}. For a limited range of image synthesis and editing scenarios, these non-parametric techniques allow non-experts to author photorealistic imagery.


In the last two years, the idea of direct image synthesis without using the traditional rendering pipeline has gotten significant interest because of promising results from deep network architectures such as Variational Autoencoders (VAEs) \cite{kingma2013auto} and Generative Adversarial Networks (GANs) \cite{goodfellow2014generative}. However, there has been little investigation of fine-grained \emph{texture} control in deep image synthesis (as opposed to coarse texture control through ``style transfer'' methods \cite{gatys2015neural}).

In this paper we introduce TextureGAN, the first deep image synthesis method which allows users to control object texture. Users ``drag'' one or more example textures onto sketched objects and the network realistically applies these textures to the indicated objects.

This ``texture fill'' operation is difficult for a deep network to learn for several reasons: (1) Existing deep networks aren't particularly good at synthesizing high-resolution texture details even without user constraints. Typical results from recent deep image synthesis methods are at low resolution (e.g. 64x64) where texture is not prominent or they are higher resolution but relatively flat (e.g. birds with sharp boundaries but few fine-scale details). (2) For TextureGAN, the network must learn to propagate textures to the relevant object boundaries -- it is undesirable to leave an object partially textured or to have the texture spill into the background. To accomplish this, the network must implicitly segment the sketched objects \emph{and} perform texture synthesis, tasks which are individually difficult. (3) The network should additionally learn to foreshorten textures as they wrap around 3d object shapes, to shade textures according to ambient occlusion and lighting direction, and to understand that some object parts (handbag clasps) are not to be textured but should occlude the texture. These texture manipulation steps go beyond traditional texture synthesis in which a texture is assumed to be stationary. To accomplish these steps the network needs a rich implicit model of the visual world that involves some partial 3d understanding.

Fortunately, the difficulty of this task is somewhat balanced by the availability of training data. Like recent unsupervised learning methods based on colorization~\cite{zhang2016colorful,larsson2016learning}, training pairs can be generated from unannotated images. In our case, input training sketches and texture suggestions are automatically extracted from real photographs which in turn serve as the ground truth for initial training. We introduce \emph{local} texture loss to further fine-tune our networks to handle diverse textures unseen on ground truth objects.

We make the following contributions:
\begin{itemize}
  \item We are the first to demonstrate the plausibility of fine-grained texture control in deep image synthesis. In concert with sketched object boundaries, this allows non-experts to author realistic visual content. Our network is feed-forward and thus can run interactively as users modify sketch or texture suggestions.
  \item We propose a ``drag and drop'' texture interface where users place particular textures onto sparse, sketched object boundaries. The deep generative network directly operates on these localized texture patches and sketched object boundaries.
    \item We explore novel losses for training deep image synthesis. In particular we formulate a local texture loss which encourages the generative network to handle new textures never seen on existing objects.

\end{itemize}




\section{Related Work}
\textbf{Image Synthesis.}
Synthesizing natural images has been one of the most intriguing and challenging tasks in graphics, vision, and machine learning research. Existing approaches can be grouped into non-parametric and parametric methods. On one hand, non-parametric approaches have a long-standing history. They are typically data-driven or example-based, i.e., directly exploit and borrow existing image pixels for the desired tasks~\cite{barnes2009patchmatch,sketch2photo,efros1999texture,hays2007scene,mcmillan1995plenoptic}. Therefore, non-parametric approaches often excel at generating realistic results while having limited generalization ability, i.e., being restricted by the limitation of data and examples, e.g., data bias and incomplete coverage of long-tail distributions. On the other hand, parametric approaches, especially deep learning based approaches, have achieved promising results in recent years. Different from non-parametric methods, these approaches utilize image datasets as training data to fit deep parametric models, and have shown superior modeling power and generalization ability in image synthesis~\cite{goodfellow2014generative,kingma2013auto}, e.g., hallucinating diverse and relatively realistic images that are different from training data.

Generative Adversarial Networks (GANs)~\cite{goodfellow2014generative} are a type of parametric method that has been widely applied and studied for image synthesis. The main idea is to train paired generator and discriminator networks jointly. The goal of the discriminator is to classify between `real' images and generated `fake' images. The generator aims to fool the discriminator by generating images which are indistinguishable from real images. Once trained, the generator can be used to synthesize images when seeded with a noise vector. Compared to the blurry and low-resolution outcome from other deep learning methods~\cite{kingma2013auto,generateChairs}, GAN-based methods~\cite{radford2015unsupervised,mao2016least,isola2016image,zhu2016generative} generate more realistic results with richer local details and of higher resolution. 

\textbf{Controllable Image Synthesis and Conditional GANs.} 
Practical image synthesis tools require human-interpretable controls. These controls could range from high-level attributes, such as object classes \cite{odena2016conditional}, object poses~\cite{generateChairs}, natural language descriptions~\cite{reed2016generative}, to fine-grained details, such as segmentation masks~\cite{isola2016image}, sketches~\cite{sangkloy2016scribbler,guccluturk2016sketchinv}, color scribbles~\cite{sangkloy2016scribbler,zhang2017real}, and cross-domain images~\cite{gatys2015neural,yoo2016pixel}.

While the `vanilla' GAN is able to generate realistic looking images from noise, it is not easily controllable. \emph{Conditional} GANs are models that synthesize images based on input modalities other than simple noise, thus offering more control over the generated results. Compared to vanilla GANs, conditional GANs introduce additional discriminators or losses to guide generators to output images with desired properties, e.g., an object category discriminator~\cite{odena2016conditional}, a discriminator to judge visual-text association~\cite{reed2016generative}, or a simple pixel-wise loss between generated images and target images~\cite{isola2016image}. 

It is worth highlighting several recent works on sketch or color-constrained deep image synthesis. Scribbler~\cite{sangkloy2016scribbler} takes as input a sketch and short color strokes, and generates realistically looking output that follows the input sketch and has color consistent with the color strokes. A similar system is employed for automatically painting cartoon images~\cite{liu2017auto}. A user-guided interactive image colorization system was proposed in~\cite{zhang2017real}, offering users the control of color when coloring or recoloring an input image. Distinct from these works, our system simultaneously supports richer user guidance signals including structural sketches, color patches, \emph{and} texture swatches. Moreover, we examine new loss functions.

\textbf{Texture Synthesis and Style Transfer.}
Texture synthesis and style transfer are two closely related topics in image synthesis. Given an input texture image, texture synthesis aims at generating new images with visually similar textures. Style transfer has two inputs -- \emph{content} and \emph{style} images -- and  aims to synthesize images with the layout and structure of the content image and the texture of the style image. 
Non-parametric texture synthesis and style transfer methods typically resample provided example images to form the output~\cite{efros1999texture,efros2001image,wei2000fast,hertzmann2001image}. TextureShop~\cite{TextureShop} is similar to our method in that it aims to texture an object with a user-provided texture, but the technical approach is quite different. TextureShop uses non-parametric texture synthesis and shape-from-shading to foreshorten the texture so that it appears to follow the surface of a photographed object. 

A recent deep style transfer method by Gatys et al.~\cite{gatys2015texture,gatys2015neural} demonstrates that the correlations (i.e., Gram matrix) between features extracted from a pre-trained deep neural network capture the characteristics of textures well and showed promising results in synthesizing textures and transferring styles. Texture synthesis and style transfer are formalized as an optimization problem, where an output image is generated by minimizing a loss function of two terms, one of which measures content similarity between the input content image and the output, and the other measures style similarity between the input style and the output using the Gram matrix. Since the introduction of this approach by Gatys et al.~\cite{gatys2015texture,gatys2015neural}, there have been many works on improving the generalization~\cite{zhang2017multi,huang2017arbitrary,li2017diversified}, efficiency~\cite{ulyanov2016texture,johnson2016perceptual} and controllability~\cite{gatys2016controlling} of deep style transfer. 

Several texture synthesis methods use GANs to improve the quality of the generated results. Li and Wand~\cite{li2016precomputed} use adversarial training to discriminate between real and fake textures based on a feature patch from the VGG network. Instead of operating on feature space, Jetchev et al.~\cite{jetchev2016texture} and Bergman et al.~\cite{bergmann2017learning} apply adversarial training at the pixel level to encourage the generated results to be indistinguishable from  real texture. Our proposed texture discriminator in Section \ref{sec:patch_based_texture_loss} differs from prior work by comparing a \emph{pair} of patches from generated and ground truth textures instead of using a single texture patch. Intuitively, our discriminator is tasked with the fine-grained question of ``is this the same texture?'' rather than the more general ``is this a valid texture?''. Fooling such a discriminator is more difficult and requires our generator to synthesize not just realistic texture but also texture that is faithful to various input texture styles.

Similar to texture synthesis, image completion or inpainting methods also show promising results using GANs. Our task has similarities to the image completion problem, which attempts to fill in missing regions of an image, although our missing area is significantly larger and partially constrained by sketch, color, or texture. Similar to our approach, Yang et al.~\cite{Yang_2017_CVPR} computes texture loss between patches to encourage the inpainted region to be faithful to the original image regions. However, their texture loss only accounts for similarity in feature space. Our approach is similar in spirit to Iizuka et al.~\cite{IizukaSIGGRAPH2017}, which proposes using both global and local discriminators to ensure that results are both realistic and consistent with the image context, whereas our local discriminator is instead checking texture similarity between input texture patch and output image.

\begin{figure}
  \centering
  \includegraphics[width=1\linewidth]{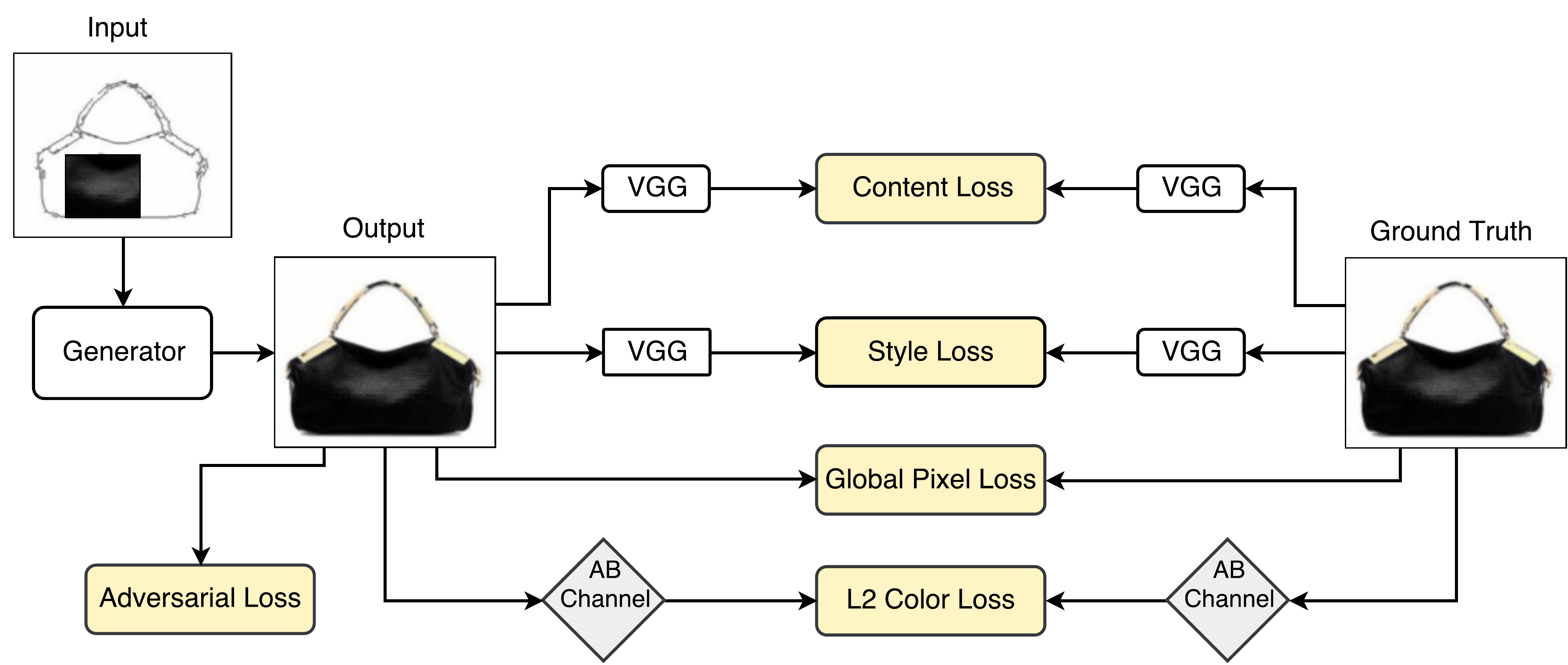}
  \caption{TextureGAN pipeline for the ground-truth pre-training (section \ref{sec:gt_training}) 
  }
  \label{fig:diagram_gt}
\end{figure}

  %
\begin{figure}
  \centering
  \includegraphics[width=1\linewidth]{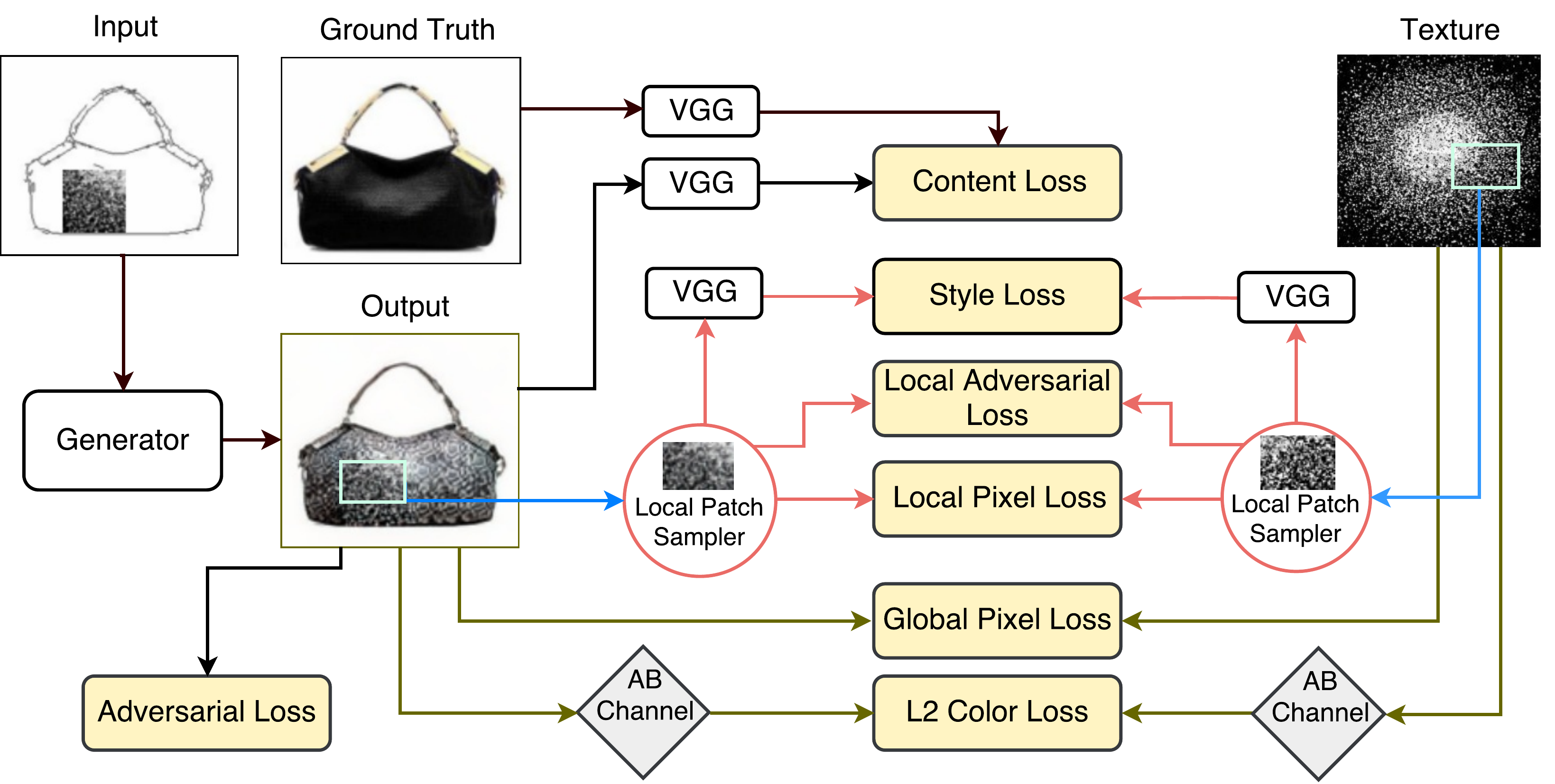}
  \caption{TextureGAN pipeline for the external texture fine-tuning (section \ref{sec:txt_training})
  }
  \label{fig:diagram}
\end{figure}

  %
\section{TextureGAN}

We seek an image synthesis pipeline that can generate natural images based on an input \emph{sketch} and some number of user-provided \emph{texture} patches. 
Users provide rough sketches that outline the desired objects to control the generation of semantic content, e.g. object type and shape, but sketches do not contain enough information to guide the generation of texture details, materials, or patterns. 
To guide the generation of fine-scale details, we want users to somehow control texture properties of objects and scene elements.
%
%

Towards this goal, we introduce \textbf{TextureGAN}, a conditional generative network that learns to generate realistic images from input sketches with overlaid textures. 
%
We argue that instead of providing an unanchored texture sample, users can more precisely control the generated appearance by directly placing small texture patches over the sketch, since locations and sizes of the patches provide hints about the object appearance desired by the user. 
%
%
In this setup, the user can `drag' rectangular texture patches of arbitrary sizes onto different sketch regions as additional input to the network. For example, the user can specify a striped texture patch for a shirt and a dotted texture patch for a skirt. The input patches guide the network to propagate the texture information to the relevant regions respecting semantic boundaries (e.g. dots should appear on the skirt but not on the legs).

A major challenge for a network learning this task is the uncertain pixel correspondence between the input texture and the unconstrained sketch regions. To encourage the network to produce realistic textures, we propose a local texture loss (Section \ref{sec:txt_training}) based on a texture discriminator and a Gram matrix style loss. 
This not only helps the generated texture follow the input faithfully, but also helps the network learn to propagate the texture patch and synthesize new texture. 

TextureGAN also allows users to more precisely control the colors in the generated result. One limitation of previous color control with GANs~\cite{sangkloy2016scribbler} is that the input color constraints in the form of~$\mathbf{RGB}$~need to fight with the network's understanding about the semantics, e.g., bags are mostly black and shoes are seldom green. 
To address this problem, we train the network to generate images in the $\mathbf{Lab}$ color space. We convert the groundtruth images to $\mathbf{Lab}$, enforce the content, texture and adversarial losses only on the $\mathbf{L}$ channel, and enforce a separate color loss on the $\mathbf{ab}$ channels. 
We show that combining the controls in this way allows the network to generate realistic photos closely following the user's color and texture intent without introducing obvious visual artifacts.
We use the network architecture proposed in Scribbler~\cite{sangkloy2016scribbler} with additional skip connections. Details of our network architecture are 
included in the supplementary material.
We use a 5-channel image as input to the network. The channels support three different types of controls -- one channel for sketch, two channels for texture (one intensity and one binary location mask), and two channels for color. 
%
%
Section~\ref{sec:training-data} describes the method we used to generate each input channel of the network.

We first train TextureGAN to reproduce ground-truth shoe, handbag, and clothes photos given synthetically sampled input control channels. 
We then generalize TextureGAN to support a broader range of textures and to propagate unseen textures better by fine-tuning the network with a separate texture-only database.

\subsection{Ground-truth Pre-training}
\label{sec:gt_training}
We aim to propagate the texture information contained in small patches to fill in an entire object. 
As in Scribbler~\cite{sangkloy2016scribbler}, we use feature and adversarial losses to encourage the generation of realistic object structures. However, we find that these losses alone cannot reproduce fine-grained texture details. 
Also, Scribbler uses pixel loss to enforce color constraints, but fails when the input color is rare for that particular object category. 
%
%
%
%
Therefore, we redefine the feature and adversarial losses and introduce new losses to improve the replication of texture details and encourage precise propagation of colors. 
For initial training, we derive the network's input channels from ground-truth photos of objects. When computing the losses, we compare the generated images with the ground-truth.
Our objective function consists of multiple terms, each of which encourages the network to focus on different image aspects. Figure~\ref{fig:diagram_gt} shows our pipeline for the ground-truth pre-training.

\textbf{Feature Loss $\mathcal{L}_F$. }
It has been shown previously that the features extracted from middle layers of a pre-trained neural network, VGG-19~\cite{simonyan2014very}, represent high-level semantic information of an image~\cite{guccluturk2016sketchinv, sangkloy2016scribbler}. Given a rough outline sketch, we would like the generated image to loosely follow the object structures specified by the sketch. Therefore, we decide to use a deeper layer of VGG-19 for feature loss (relu 4\_2). To focus the feature loss on generating structures, we convert both the ground-truth image and the generated image from $\mathbf{RGB}$ color space to $\mathbf{Lab}$ and generate grayscale images by repeating the $\mathbf{L}$ channel values. We then feed the grayscale image to VGG-19 to extract features. The feature loss is defined as the L2 difference in the feature space. During back propagation, the gradients passing through the L channel of the output image are averaged from the three channels of the VGG-19 output. 
\begin{figure}
\centering
\includegraphics[width=0.9\linewidth]{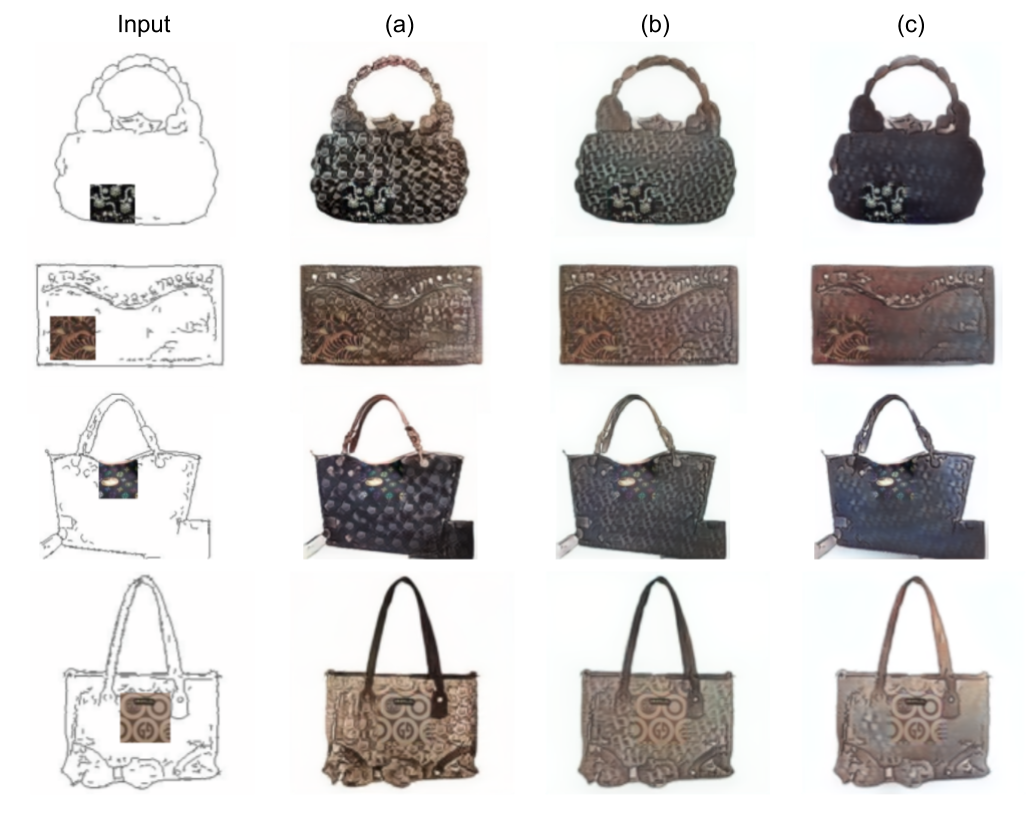}
\caption{The effect of texture loss and adversarial loss. a) The network trained using all proposed losses can effectively propagate textures to most of the foreground region; b) Removing adversarial loss leads to blurry results; c) Removing texture loss harms the propagation of textures.  }
\label{fig:ablation-study}
\end{figure}

\textbf{Adversarial Loss $\mathcal{L}_{ADV}$. }
%
%
%
In recent work, the concept of adversarial training has been adopted in the context of image to image translation. In particular, one can attach a trainable discriminator network at the end of the image translation network and use it to constrain the generated result to lie on the training image manifold. Previous work proposed to minimize the adversarial loss (loss from the discriminator network) together with other standard losses (pixel, feature losses, etc). The exact choice of losses depends on the different applications~\cite{sangkloy2016scribbler, isola2016image, guccluturk2016sketchinv}. 
Along these lines, we use adversarial loss on top of feature, texture and color losses. The adversarial loss pushes the network towards synthesizing sharp and realistic images, but at the same time constrains the generated images to choose among typical colors in the training images. The network's understanding about color sometimes conflicts with user's color constraints, e.g. a user provides a rainbow color constraint for a handbag, but the adversarial network thinks it looks fake and discourages the generator from producing such output. 
Therefore, we propose applying the adversarial loss $\mathcal{L}_{adv}$ only on grayscale image (the $\mathbf{L}$ channel in $\mathbf{Lab}$ space). The discriminator is trained to disregard the color but focus on generating sharp and realistic details. The gradients of the loss only flow through the $\mathbf{L}$ channel of the generator output. This effectively reduces the search space and makes GAN training easier and more stable. 
We perform the adversarial training using the techniques proposed in DCGAN~\cite{radford2015unsupervised} with the modification proposed in LSGAN~\cite{mao2016least}. LSGAN proposed replacing the cross entropy loss in the original GAN with least square loss for higher quality results and stable training.


\textbf{Style Loss $\mathcal{L}_S$. }
In addition to generating the right content following the input sketch, we would also like to propagate the texture details given in the input texture patch. The previous feature and adversarial losses sometimes struggle to capture fine-scale details, since they focus on getting the overall structure correct. 
Similar to deep learning based texture synthesis and style transfer work~\cite{gatys2015texture, gatys2015neural}, we use style loss to specifically encourage the reproduction of texture details, but we apply style loss on the $\mathbf{L}$ channel only. We adopt the idea of matching the Gram matrices (feature correlations) of the features extracted from certain layers of the pretrained classification network (VGG-19). The Gram matrix $\mathcal{G}^{l}_{ij} \in 
\mathcal{R}^{{N}_{l}\times{N}_{l}}$ is defined as:
\begin{equation}\label{eq:gram-matrix}
\mathcal{G}^{l}_{ij} = \sum_{k}\mathcal{F}^{l}_{ik}\mathcal{F}^{l}_{jk}
\end{equation}
where, $N_l$ is the number of feature maps at network layer $l$, $\mathcal{F}^{l}_{ik}$ is the activation of the $i$th filter at position $k$ in layer $l$. 
We use two layers of the VGG-19 network (relu3\_2, relu4\_2) to define our style loss.

\textbf{Pixel Loss $\mathcal{L}_P$. }
We find that adding relatively weak L2 pixel loss on the $\mathbf{L}$ channel stabilizes the training and leads to the generation of texture details that are more faithful to the user's input texture patch. 



\textbf{Color Loss $\mathcal{L}_C$. }
All losses above are applied only on the $\mathbf{L}$ channel of the output to focus on generating sketch-conforming structures, realistic shading, and sharp high-frequency texture details. To enforce the user's color constraints, we add a separate color loss that penalizes the L2 difference between the $\mathbf{ab}$ channels of the generated result and that of the ground-truth. 



Our combined objective function is defined as:
\begin{equation}\label{eq:objective-function}
\mathcal{L} = \mathcal{L}_{F} + \mathbf{w}_{ADV}\mathcal{L}_{ADV} + \mathbf{w}_{S}\mathcal{L}_{S} + 
\mathbf{w}_{P}\mathcal{L}_{P} + \mathbf{w}_{C}\mathcal{L}_{C}
\end{equation}

\subsection{External Texture Fine-tuning}
\label{sec:txt_training}

One problem of training with ``ground-truth'' images is that it is hard for the network to focus on reproducing low-level texture details due to the difficulty of disentangling the texture from the content within the same image. For example, we do not necessarily have training examples of the same object with different textures applied which might help the network learn the factorization between structure and texture.
Also, the Gram matrix-based style loss can be dominated by the feature loss since both are optimized for the same image. There is not much room for the network to be creative in hallucinating low-level texture details, since it tends to focus on generating high-level structure, color, and patterns. 
Finally, many of the ground-truth texture patches contain smooth color gradients without rich details. Trained solely on those, the network is likely to ignore ``hints'' from an unseen input texture patch at test time, especially if the texture hint conflicts with information from the sketch. 
%
As a result, the network often struggles to propagate high-frequency texture details in the results especially for textures that are rarely seen during training. 

To train the network to propagate a broader range of textures, we fine-tune our network to reproduce and propagate textures \emph{for which we have no ground truth output}. To do this, we introduce a new local texture loss and adapt our existing losses to encourage faithfulness to a \emph{texture} rather than faithfulness to a ground truth output \emph{object photo}. 
%
We use all the losses introduced in the previous sections except the global style loss $\mathcal{L}_{S}$. We keep the \textit{feature and adversarial losses}, $\mathcal{L}_{F}, \mathcal{L}_{ADV}$, unchanged, but modify the \textit{pixel and color losses}, $\mathcal{L}^{\prime}_{P}, \mathcal{L}^{\prime}_{C}$, to compare the generated result with the entire input texture from which input texture patches are extracted. Figure~\ref{fig:diagram} shows our pipeline for the external texture fine-tuning. 
To prevent color and texture bleeding, the losses are applied only on the foreground object, as approximated by a segmentation mask (Section~\ref{sec:segmentation-mask}).

\begin{figure}
  \centering
  \includegraphics[width=\linewidth]{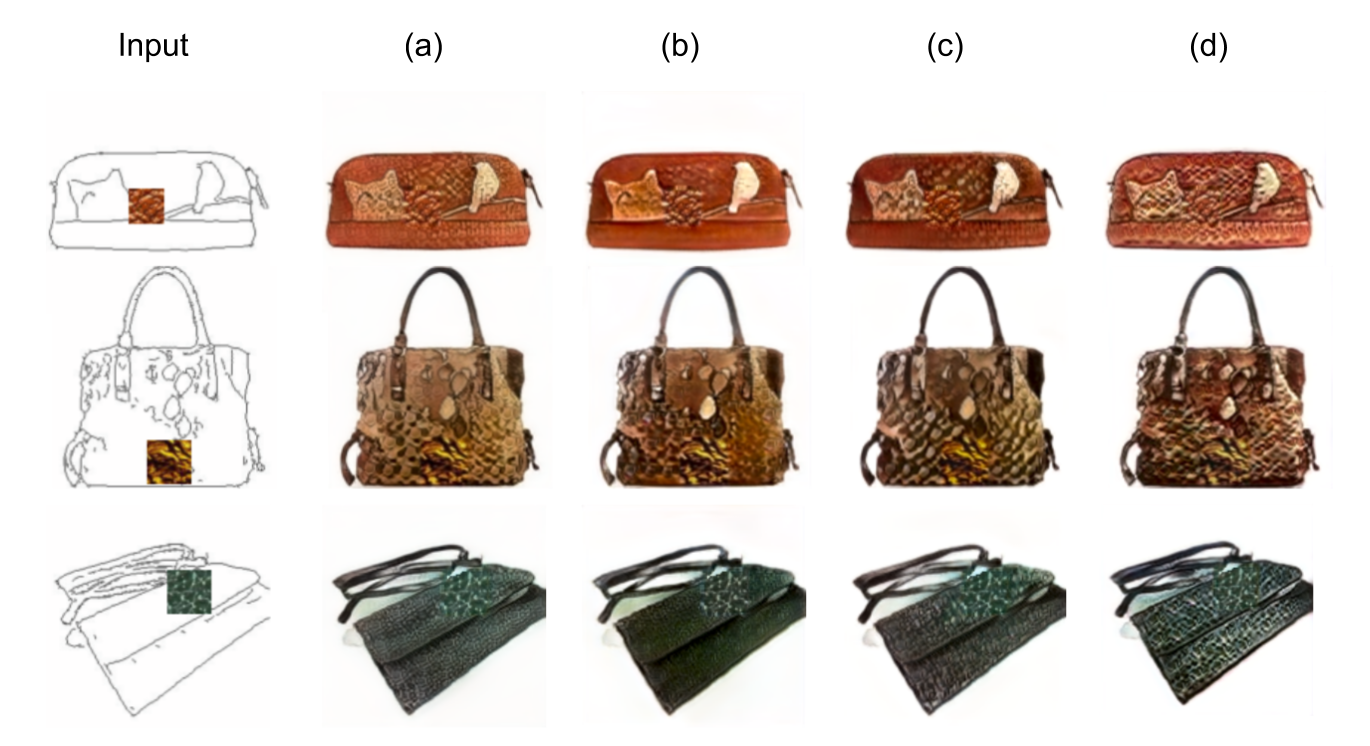}
  \caption{Effect of proposed \emph{local} texture losses. Results from the ground-truth model a) without any local losses, b) with local pixel loss, c) with local style loss, d) with local adversarial loss. 
With local adversarial loss, the network tends to produce more consistent texture throughout the object. 
}
  \label{fig:dtd-training-effect}
\end{figure}

\subsubsection{Local Texture Loss}
\label{sec:patch_based_texture_loss}
To encourage better propagation of texture, we propose a \textbf{local texture loss $\mathcal{L}_{t}$}, that is only applied to small local regions of the output image. We randomly sample $n$ patches of size $s \times s$ from the generated result and the input texture $I_t$ from a separate texture database.
%
We only sample patches which fall inside an estimated foreground segmentation mask $R$ (section~\ref{sec:segmentation-mask}). 
%
%
The local texture loss $\mathcal{L}_{t}$ is composed of three terms:
\begin{equation}
\mathcal{L}_t = \mathcal{L}_s + \mathbf{w}_p\mathcal{L}_p + \mathbf{w}_{adv}\mathcal{L}_{adv}
\end{equation}

\textbf{Local Adversarial Loss $\mathcal{L}_{adv}$. } We introduce a local adversarial loss that decides whether a pair of texture patches have the same textures. We train a local texture discriminator $\mathbf{D_{txt}}$ to recognize a pair of cropped patches from the same texture as a positive example ($D_{txt} (\cdot) = 1$), and a pair of patches from different textures as a negative example ($D_{txt} (\cdot) = 0$). 

Let $h(x,R)$ be a cropped patch of size $s \times s$ from image $x$ based on segmentation mask $R$. Given a pair of cropped patches
$(PG_{i},PT_{i}) = (h(\mathbf{G}(x_i),R_i),h(I_t,R_i))$, 
we define $L_{adv}$ as follows: 
\begin{equation}
\mathcal{L}_{adv} = -\sum_i{(\mathbf{D_{txt}}(PG_{i},PT_{i})-1)^2}
\end{equation}


\textbf{Local Style Loss $\mathcal{L}_s$ and Pixel Loss $\mathcal{L}_p$. }
To strengthen the texture propagation, we also use Gram matrix-based style loss and L2 pixel loss on the cropped patches. \\
\\
%
While performing the texture fine-tuning, the network is trying to adapt itself to understand and propagate new types of textures, and might `forget' what it learnt from the ground-truth pretraining stage. Therefore, when training on external textures, we mix in iterations of ground-truth training fifty percent of the time.\\ 

Our final objective function becomes:
\begin{equation}\label{eq:objective-function}
\mathcal{L} = \mathcal{L}_{F} + \mathbf{w}_{ADV}\mathcal{L}_{ADV} + \mathbf{w}_{P}\mathcal{L}^\prime_{P} + \mathbf{w}_{C}\mathcal{L}^\prime_{C} + \mathcal{L}_{t}
\end{equation}

\section{Training Setup}
We train TextureGAN on three object-centric datasets -- \textbf{handbags}~\cite{zhu2016generative}, \textbf{shoes}~\cite{yu2014fine} and \textbf{clothes}~\cite{ATR,liang2015human,liuLQWTcvpr16DeepFashion,liuYLWTeccv16FashionLandmark}. 
Each photo collection contains large variations of colors, materials, and patterns. These domains are also chosen so that we can demonstrate plausible product design applications.
For supervised training, we need to generate (input, output) image pairs. For the output of the network, we convert the ground-truth photos to $\mathbf{Lab}$ color space. For the input to the network, we process the ground-truth photos to extract 5-channel images. The five channels include one channel for the binary sketch, two channels for the texture (intensities and binary location masks), and two channels for the color controls.

In this section, we describe how we obtain segmentation masks used during training, how we generate each of the input channels for the ground-truth pre-training, and how we utilize the separate texture database for the network fine-tuning. We also provide detailed training procedures and parameters.

\subsection{Segmentation Mask}
\label{sec:segmentation-mask}
For our local texture loss, we hope to encourage samples of output texture to match samples of input texture. But the output texture is localized to particular image regions (e.g. the interior of objects) so we wouldn't want to compare a background patch to an input texture. Therefore we only sample patches from within the foreground. Our handbag and shoe datasets are product images with consistent, white backgrounds so we simply set the white pixels as background pixels. For clothes, the segmentation mask is already given in the dataset ~\cite{Lassner:GeneratingPeople:2017, liang2015human}. With the clothes segmentation mask, we process the ground-truth photos to white out the background. Note that segmentation masks are \emph{not used} at test time.

\subsection{Data Generation for Pre-training}
\label{sec:training-data}

\textbf{Sketch Generation. }
For handbags and shoes, we generate sketches using the deep edge detection method used in pix2pix~\cite{xie15hed, isola2016image}.
For clothes, we leverage the clothes parsing information provided in the dataset~\cite{ATR,liang2015human}. We apply Canny edge detection on the clothing segmentation mask to extract the segment boundaries and treat them as a sketch. We also apply xDoG \cite{winnemoller2012xdog} on the clothes image to obtain more variation in the training sketches. Finally, we mix in additional synthetic sketches generated using the methods proposed in Scribbler~\cite{sangkloy2016scribbler}.

%

\textbf{Texture Patches. }
To generate input texture constraints, we randomly crop small regions within the foreground objects of the ground-truth images. 
We randomly choose the patch location from within the segmentation and randomize the patch size.
%
We convert each texture patch to the $\mathbf{Lab}$ color space and normalize the pixels to fall into 0-1 range. 
For each image, we randomly generate one or two texture patches. For clothes, we extract texture patches from one of the following regions -- top, skirt, pant, dress, or bag. We compute a binary mask to encode the texture patch location. 

\subsection{Data Generation for Fine-tuning}
To encourage diverse and faithful texture reproduction, we fine-tune TextureGAN by applying external texture patches from a leather-like texture dataset. We queried ``leather'' in Google and manually filtered the results to  130 high resolution leather textures. From this clean dataset, we sampled roughly 50 crops of size 256x256 from each image to generate a dataset of 6,300 leather-like textures. 
We train our models on leather-like textures since they are commonly seen materials for handbags, shoes and clothes and contain large appearance variations that are challenging for the network to propagate. 


\subsection{Training Details}

\label{sec:training-details}
For \textbf{pre-training}, we use the following parameters on all datasets. $\mathbf{w}_{ADV} = 1$, $\mathbf{w}_{S} = 0.1$, $\mathbf{w}_P = 10$ and $\mathbf{w}_{C} = 100$. We use the Adam optimizer~\cite{kingma2014adam} with learning rate 1e-2.

For \textbf{fine-tuning}, we optimize all the losses at the same time but use different weight settings. $\mathbf{w}_{ADV} = 1e4$, $\mathbf{w}_{S} = 0$, $\mathbf{w}_P = 1e2$, $\mathbf{w}_{C} = 1e3$, $\mathbf{w}_s = 10$, $\mathbf{w}_p = 0.01$, and $\mathbf{w}_{adv} = 7e3$.
We also decrease the learning rate to 1e-3. 
We train most of the models at input resolution of 128x128 except one clothes model at the resolution of 256x256 (Figure~\ref{fig:results-cloths}).


\section{Results and Discussions}
\textbf{Ablation Study. }
Keeping other settings the same, we train networks using different combinations of losses to analyze how they influence the result quality. In Figure~\ref{fig:ablation-study}, given the input sketch, texture patch and color patch (first column), the network trained with the complete objective function (second column) correctly propagates the color and texture to the entire handbag. If we turn off the texture loss (fourth column), the texture details within the area of the input patch are preserved, but difficult textures cannot be fully propagated to the rest of the bag. If we turn off the adversarial loss (third column), texture is synthesized, but that texture is not consistent with the input texture. Our ablation experiment confirms that style loss alone is not sufficient to encourage texture propagation motivating our local patch-based texture loss (Section~\ref{sec:patch_based_texture_loss}).
\begin{figure}
  \centering
  \includegraphics[width=\linewidth]{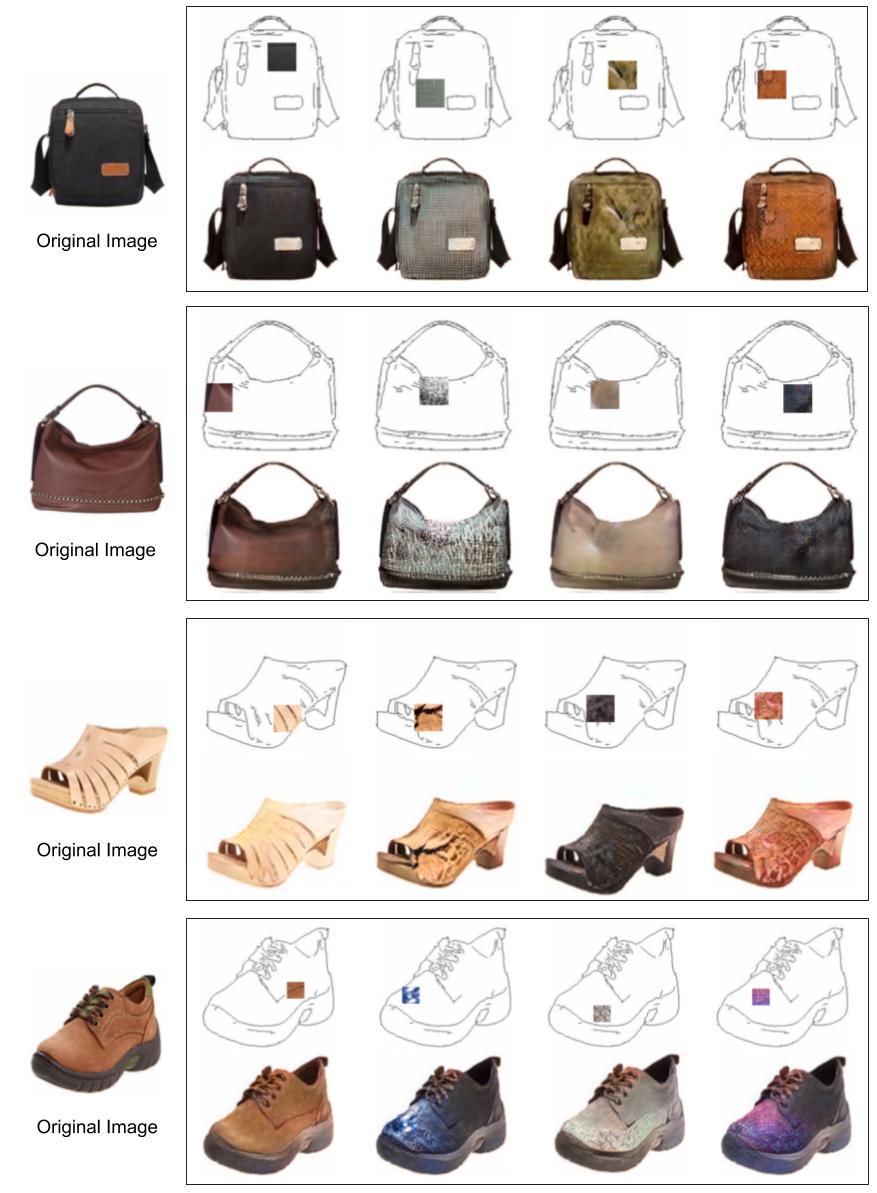}
  \caption{Results on held out shoes and handbags sketches [152x152]. On the far left is the ``ground truth'' photo from which the sketch was synthesized. On the first result column, a texture patch is also sampled from the original shoe. We show three additional results with diverse textures.
}
  \vspace{-5mm}
  \label{fig:results_gt}
\end{figure}

\begin{figure*}
  \centering
  \includegraphics[width=\linewidth]{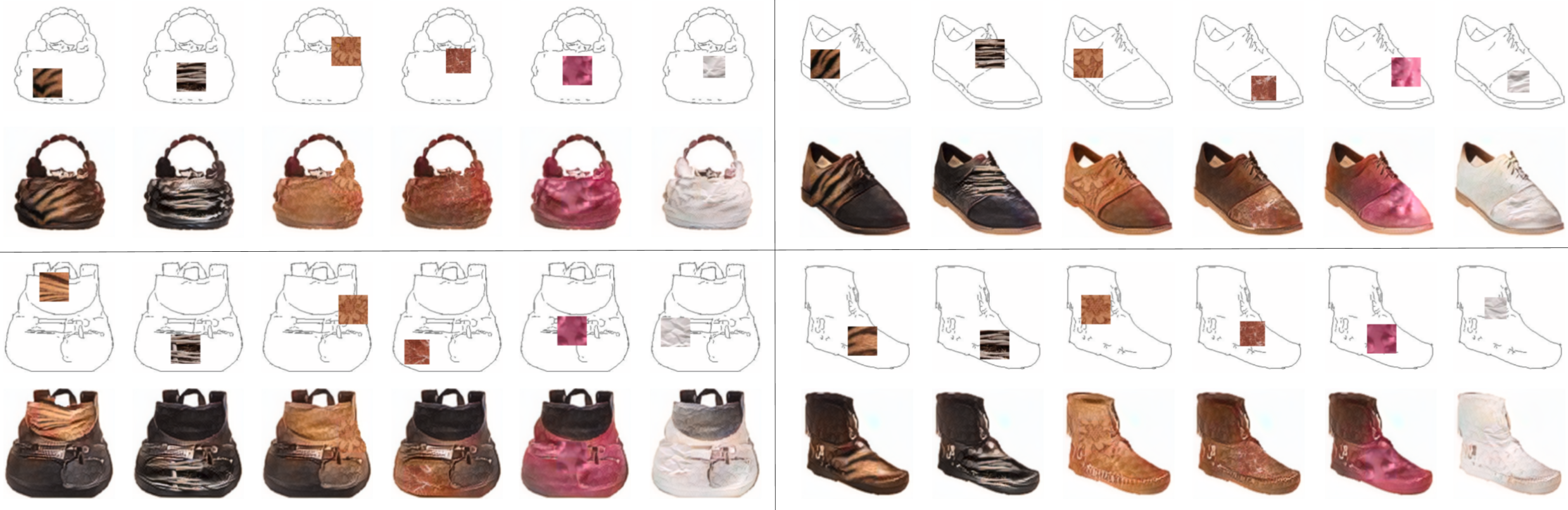}
  \caption{Results for shoes and handbags on different textures. Odd rows: input sketch and texture patch. Even rows: generated results.}
  \vspace{-6mm}
  \label{fig:results-handbag}
\end{figure*}
\textbf{External Texture Fine-tuning Results. }
We train TextureGAN on three datasets -- shoes, handbags, and clothes -- with increasing levels of structure complexity. We notice that for object categories like shoes that contain limited structure variations, the network is able to quickly generate realistic shading and structures and focus its remaining capacity for propagating textures. The texture propagation on the shoes dataset works well even without external texture fine-tuning. 
For more sophisticated datasets like handbags and clothes, external texture fine-tuning is critical for the propagation of difficult textures that contain sharp regular structures, such as stripes. 

Figure~\ref{fig:dtd-training-effect} demonstrates how external texture fine-tuning with our proposed texture loss can improve the texture consistency and propagation. 

The ``ground truth'' pre-trained model is faithful to the input texture patch in the output only directly under the patch and does not propagate it throughout the foreground region. 
By fine-tuning the network with texture examples and enforcing local style loss, local pixel loss, and local texture loss we nudge the network to 
apply texture consistently across the object.  
With local style loss (column c) and local texture discriminator loss (column d), the networks are able to propagate texture better than without fine-tuning (column a) or just local pixel loss (column b). Using local texture discriminator loss tends to produce more visually similar result to the input texture than style loss.

\begin{figure}
  \centering
  \includegraphics[width=0.9\linewidth]{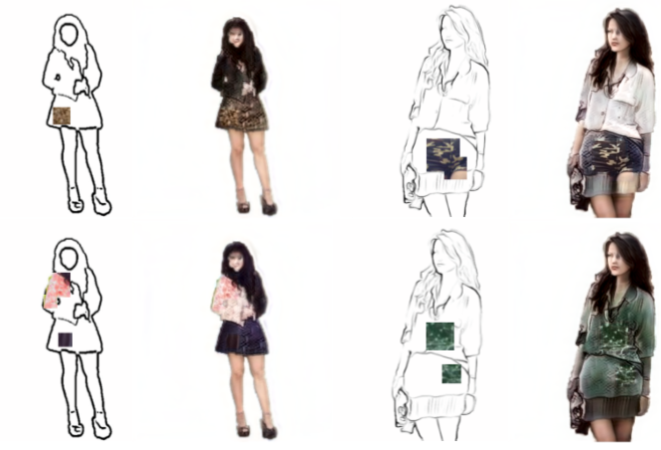}
  \caption{Applying multiple texture patches on the sketch. Our system can also handle multiple texture inputs and our network can follow sketch contours and expand the texture to cover the sketched object.}
  \label{fig:results-cloths}
\end{figure}
\begin{figure}
  \centering
  \includegraphics[width=0.8\linewidth]{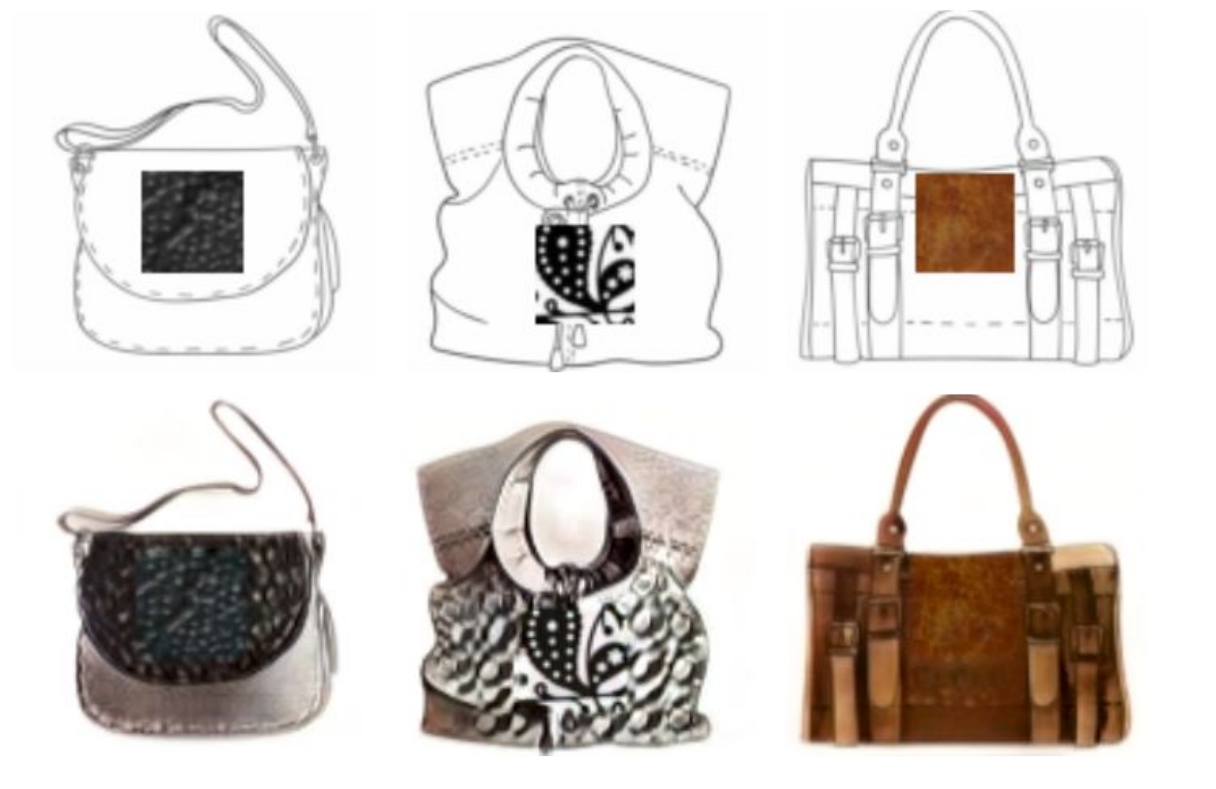}
  \caption{Results on human-drawn sketches. Sketch images from olesiaagudova - stock.adobe.com}
  \vspace{-5mm}
  \label{fig:human-sketch}
\end{figure}





Figures~\ref{fig:results_gt} and ~\ref{fig:results-handbag} show the results of applying various texture patches to sketches of handbags and shoes. These results are typical of test-time result quality. The texture elements in the camera-facing center of the bags tend to be larger than those around the boundary. Textures at the bottom of the objects are often shaded darker than the rest, consistent with top lighting or ambient occlusion. Note that even when the input patch goes out of the sketch boundary, the generated texture follow the boundary exactly.

Figure~\ref{fig:results-cloths} shows results on the clothes dataset trained at a resolution of 256x256. The clothes dataset contains large variations of structures and textures, and each image in the dataset contains multiple semantic regions. Our network can handle multiple texture patches placed on different parts of the clothes (bottom left). The network can propagate the textures within semantic regions of the sketch while respecting the sketch boundaries.
 
Figure~\ref{fig:human-sketch} shows results on human-drawn handbags. These drawings differ from our synthetically generated training sketches but the results are still high quality.  

\vspace{-1mm}
\section{Conclusion}
We have presented an approach for controlling deep image synthesis with input sketch and texture patches. With this system, a user can sketch the object structure and precisely control the generated details with texture patches. TextureGAN is feed-forward which allows users to see the effect of their edits in real time. By training TextureGAN with local texture constraints, we demonstrate its effectiveness on sketch and texture-based image synthesis. TextureGAN also operates in {\bf Lab} color space, which enables separate controls on color and content. Furthermore, our results on fashion datasets show that our pipeline is able to handle a wide variety of texture inputs and generates texture compositions that follow the sketched contours. In the future, we hope to apply our network on more complex scenes.
\section*{Acknowledgments}
This work is supported by a Royal Thai Government Scholarship to Patsorn Sangkloy and NSF award 1561968. 

{\small
\bibliographystyle{ieee}
\bibliography{egbib}
}

\end{document}